%% file: acl2021.tex
\documentclass[11pt,a4paper]{article}
\usepackage[hyperref]{acl2021}
\usepackage{times}
\usepackage{latexsym}
\usepackage{comment}

\usepackage{microtype}

\aclfinalcopy % Uncomment this line for the final submission
\def\aclpaperid{2063} %  Enter the acl Paper ID here

\usepackage{bm}
\usepackage[ruled,vlined]{algorithm2e}
\usepackage{algorithmic}
\usepackage{bbm}
\usepackage{graphicx}
\usepackage{amsmath}
\newcommand{\tref}[1]{Table~\ref{#1}}
\newcommand{\eref}[1]{Equation~(\ref{#1})}
\newcommand{\fref}[1]{Figure~\ref{#1}}
\newcommand{\ssref}[1]{Section~\ref{#1}}

\newcommand{\appendixref}[1]{Appendix~\ref{#1}}

\newcommand{\algref}[1]{Algorithm~\ref{#1}}

\newcommand{\argmax}{\mathop{\rm arg~max}\limits}

\def\edit{\textsc{Edit}}
\def\rulebased{\textsc{Rule}}
\def\withoutgcn{\textsc{Edit-GCN}}
\def\withoutitr{\textsc{Edit-IE}}
\def\randedit{\textsc{Random Edit}}
\def\randinit{\textsc{Random Init}} 

\newcommand{\entitylabel}[1]{\textsc{#1}}
\newcommand{\relationlabel}[1]{\textsc{#1}}
\newcommand{\entity}[1]{\textit{#1}}

\title{A Neural Edge-Editing Approach for Document-Level Relation Graph Extraction}

\author{Kohei Makino \\
  \And
  Makoto Miwa \\
  Toyota Technological Institute\\
  2-12-1 Hisakata, Tempaku-ku, Nagoya, 468-8511, Japan\\
  \texttt{\{sd21505, makoto-miwa, yutaka.sasaki\}@toyota-ti.ac.jp} \\
  \And
  Yutaka Sasaki
  }

\date{}

\begin{document}
\maketitle
\begin{abstract}
In this paper, we propose a novel edge-editing approach to extract relation information from a document. We treat the relations in a document as a relation graph among entities in this approach. The relation graph is iteratively constructed by editing edges of an initial graph, which might be a graph extracted by another system or an empty graph. The way to edit edges is to classify them in a close-first manner using the document and temporally-constructed graph information; 
each edge is represented with a document context information by a pretrained transformer model and a graph context information by a graph convolutional neural network model. We evaluate our approach on the task to extract material synthesis procedures from materials science texts. The experimental results show the effectiveness of our approach in editing the graphs initialized by our in-house rule-based system and empty graphs.\footnote{The source code is available at \url{https://github.com/tti-coin/edge-editing}.}
\end{abstract}

\section{Introduction}
\label{sec:intro}
Relation extraction (RE), the task to predict relations between pairs of given entities from literature, is an important task in natural language processing. While most existing work focused on sentence-level RE~\cite{zeng-etal-2014-relation}, recent studies extended the extraction to the document level since many relations are expressed across sentences~\cite{christopoulou-etal-2019-connecting,nan-etal-2020-reasoning-lsr}. 

In document-level RE, models need to deal with relations among multiple entities over a document. Several document-level RE methods construct a document-level graph, which is built on nodes of words or other linguistic units, to capture document-level interactions between entities~\cite{christopoulou-etal-2019-connecting,nan-etal-2020-reasoning-lsr}.
However, such methods do not directly consider interactions among relations in a document, while such relations are often dependent on each other, and other relations can be considered as important contexts for a relation. 

\begin{figure}[t]
    \centering
    \includegraphics[width=\linewidth]{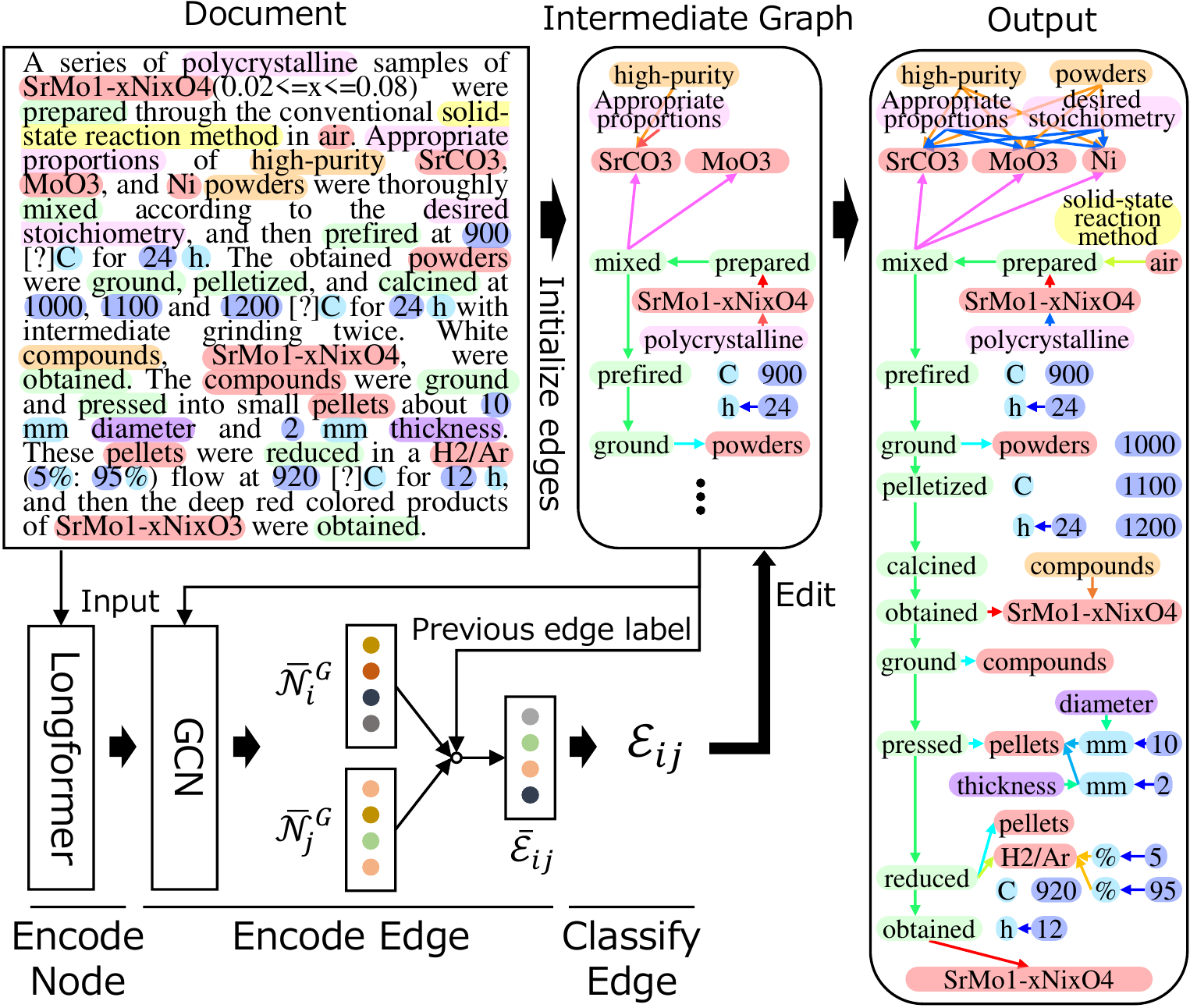}
    \caption{Overview of edge editing approach}
    \label{fig:overview}
\end{figure}

We propose a novel, iterative, edge-editing approach to document-level RE. The overview of our approach and an example of the extraction results are illustrated in \fref{fig:overview}. Our approach treats relations as a relation graph that is composed of entities as nodes and their relations as edges. 
The relation graph is first initialized using the edges predicted by an existing RE model if provided. 
Edges are then edited by a neural edge classifier that represents edges using the document information, prebuilt graph information, and the current edge information. The document information is represented with pretrained Longformer models~\cite{Beltagy2020Longformer}, while the graph information is represented with graph convolutional networks~\cite{kipf2017gcn}. Edges are edited iteratively in a close-first manner so that the approach can utilize the information of edges between close entity pairs in editing edges of distant entity pairs, which are often difficult to predict. We evaluate our approach on the task to extract synthesis procedures from text~\cite{mysore-et-al-2019-olivetti-corpus} and show the effectiveness of our approach.

The contribution of this paper is three-fold. First, we propose a novel edge-editing approach for document-level RE that utilizes contexts in both relation graphs and documents. Second, we build a strong rule-based model and show that our approach can effectively utilize and enhance the output of the rule-based model. 
Third, we build and evaluate a neural model for extracting synthesis procedures from text for the first time. 

\section{Approach}
\label{sec:method}
Our approach extracts a relation graph on given entities from a document. 
We formulate the extraction task as an edge-editing task, where the approach iteratively edits edges with a neural edge classifier in a close-first manner~\cite{miwa-sasaki-2014-modeling-joint-relation-close-first}.

\subsection{Iterative Edge Editing}
We build a relation graph by editing the edges iteratively using the edge classifier in \ssref{ssec:model}. The building finishes when all edges are edited. The edges are edited in a close-first manner~\cite{miwa-sasaki-2014-modeling-joint-relation-close-first,ma-etal-2019-easy-first} that edits the close edges first and far edges later. The distance between the entity pair is defined based on the appearing order of entities in a document; if two entities in a pair appear $m$-th and $m+3$-th, the distance becomes $3$. Note that each edge is edited only once throughout the entire editing process.

\algref{alg:edit} shows the method to build the graph by the iterative edge editing. 
To reduce the computational cost, the pairs with the same distance are edited simultaneously and the pairs with distances more than or equal to the maximum distance $d_{max}$ are edited simultaneously. 
This reduces the number of edits from $|\mathcal{N}|^2$ to $d_{max}$.

\label{ssec:edit}
\begin{algorithm}[t]
    \caption{Iterative Edge Editing}
    \label{alg:edit}
    \begin{algorithmic}
        \STATE $\mathrm{Distance}({\mathcal{N}}, d_1, d_2)$ returns pairs that have distance $d$ ($d_1 \leq d < d_2)$.
        \REQUIRE $\mathrm{doc}$: document, $\mathcal{E}$: initial edges
        \STATE $d_{max}$: maximum distance
        \ENSURE $\mathcal{E}$: edited edges
        \STATE $\bar{\mathcal{N}} \Leftarrow \mathrm{EncodeNode}(\mathrm{doc}, \mathcal{N})$
        \WHILE{$d~\mathrm{in}~\mathrm{range}(\max (|{\mathcal{N}}|, d_{max}))$}
        \STATE $\bar{\mathcal{N}}^{G} \Leftarrow \mathrm{GCN}(\bar{\mathcal{N}}, \mathcal{E})$
        \STATE $\bar{\mathcal{E}} \Leftarrow \mathrm{EncodeEdge}(\bar{\mathcal{N}}^{G},\mathcal{E})$
        \IF{$d = d_{max}$}
        \STATE $\mathcal{P} \Leftarrow \mathrm{Distance}(\mathcal{N}, d_{max}, \infty)$
        \ELSE
        \STATE $\mathcal{P} \Leftarrow \mathrm{Distance}(\mathcal{N}, d, d+1)$
        \ENDIF
        \WHILE{$(i, j) ~\mathrm{in}~\mathcal{P}$}
        \STATE $\mathcal{E}_{ij} \Leftarrow \mathrm{ClassifyEdge}(\bar{\mathcal{E}}_{ij})$
        \ENDWHILE
        \ENDWHILE
    \end{algorithmic}
\end{algorithm}

\subsection{Edge Classifier}
\label{ssec:model}

An edge classifier predicts the class of the target edge $\hat{\mathcal{E}}_{ij}$ from inputs that are composed of a document information $\mathrm{doc}$, a graph of nodes $\mathcal{N}$ and edges $\mathcal{E}$, and the node pair $(\mathcal{N}_i,\mathcal{N}_j)$ of a target edge. The classifier composed of three modules:\\ $\mathrm{\textbf{EncodeNode}}$ that produces document-based node representations $\bar{\mathcal{N}}$ using the document $\mathrm{doc}$ and the entity information of the nodes $\mathcal{N}$.\\
$\mathrm{\textbf{EncodeEdge}}$ that obtains the representation of edges $\bm \bar{\mathcal{E}}$ that applies GCN on a prebuilt graph with the node representations $\bar{\mathcal{N}}$ and edges $\mathcal{E}$.\\
$\mathrm{\textbf{ClassifyEdge}}$ that predicts the class of the edge $\hat{\mathcal{E}}_{ij}$ using the edge representation $\bar{\mathcal{E}}_{ij}$ between the node pair $(\mathcal{N}_i,\mathcal{N}_j)$.\\
We explain the details of these modules in the remaining part of this section.

$\mathrm{EncodeNode}$ employs Longformer~\cite{Beltagy2020Longformer} to obtain the document-level representation. It aggregates subword representations within each entity by max-pooling $\mathrm{Pool}$ and concatenates the aggregated information with the entity's class label representation $\bm v^{lab}$.
\begin{eqnarray}
    \bar{\mathcal{N}} &=& \mathrm{EncodeNode}(\mathrm{doc},\mathcal{N})  \nonumber \\
    &=& [\mathrm{Pool}(\mathrm{Longformer}(\mathrm{doc})); \bm v^{lab}],
\end{eqnarray}
where $[\cdot;\cdot]$ denotes concatenation.

To prepare the input to $\mathrm{EncodeEdge}$, the obtained document-based node representation is enriched by GCN to introduce the context of each node in the prebuilt graph: $\bar{\mathcal{N}}^{G} = \mathrm{GCN}(\bar{\mathcal{N}}, \mathcal{E})$.
We add inverse directions to the graph and assign different weights to different classes in graph convolutional network (GCN) following \citet{schlichtkrull-2018-r-gcn}. The produced node representation $\bar{\mathcal{N}}^{G}$ includes both document and prebuilt graph contexts.

$\mathrm{EncodeEdge}$ produces the edge representation $\bar{\mathcal{E}}$ from $\bar{\mathcal{N}}^{G}$. It individually calculates the representation of the edge $\bar{\mathcal{E}}_{ij}$ for each pair of nodes $(\mathcal{N}_i,\mathcal{N}_j)$ by combining the representations of nodes similarly to \citet{zhou-2021-atlop-relation-cdr-docred} with the embedding of the distance of the entity pair $\bm b_{ij}$ and the edge class $\bm e_{ij}^{old}$ before editing.
The distance between the entity pairs is calculated in the same way as in \ssref{ssec:edit}. If the distance exceeds a predefined maximum distance, it will be treated as the maximum distance. 
We prepare fully connected (FC) layers, $\mathrm{FC}^{H}$ and $\mathrm{FC}^{T}$, for the start point (head) and end point (tail) nodes and calculate the edge representation as follows:
\begin{eqnarray}
    \bar{\mathcal{E}}_{ij}=& \mathrm{EncodeEdge}(\bar{\mathcal{N}}^{G},\mathcal{E})_{ij} \nonumber\\
 =& [\mathrm{FC}^{H}(\bar{\mathcal{N}}_i^{G})^\top \bm W \mathrm{FC}^{T}( \bar{\mathcal{N}}_j^{G}); \bm b_{ij}; \bm e_{ij}^{old}],\label{eq:edgeij}
\end{eqnarray}
where $\bm W$ denotes a trainable weight parameter.

$\mathrm{ClassifyEdge}$ classifies the target edge $\mathcal{E}_{ij}$ into a relation class or no relation. It applies a dropout layer~\cite{srivastava-2014-dropout}, a FC layer for output $\mathrm{FC}^{out}$ and softmax to the edge representation $\bar{\mathcal{E}}_{ij}$ to predict the class $\hat{\mathcal{E}}_{ij}$ with the highest probability. 
\begin{eqnarray}
    {\hat{\mathcal{E}}}_{ij} =& \mathrm{ClassifyEdge}(\bm \bar{\mathcal{E}}_{ij}) =  \argmax \bm{\hat p}_{ij} \nonumber\\
    \bm{\hat p}_{ij} =& \mathrm{Softmax}(\mathrm{FC}^{out}(\mathrm{Dropout}(\bar{\mathcal{E}}_{ij}))) 
\end{eqnarray}

We maximize the log-likelihood in training the edge classifier. 

\section{Experiments}

\subsection{Experimental Settings}

We evaluate our approach on the materials science procedural text corpus~\cite{mysore-et-al-2019-olivetti-corpus}. In the corpus, the synthesis procedures are annotated as a graph in a document, where 19 node types such as materials, operations, and conditions and 15 directed relation types are defined. The corpus consists of 200 documents for training, 15 for development, and 15 for test. The statistics of the corpus are shown in \appendixref{sec:stat-corpus}. We chose this corpus since this corpus is publicly available, manually annotated, and it deals with a dense document-level relation graph.

We prepared a rule-based model (\textbf{\rulebased{}}) as a baseline and as an existing model to initialize the edges, which was adapted from the rule-based system in \citet{kuniyoshi-etal-2020-annotating}. The rules are summarized in \appendixref{sec:rule}. 

We employ the micro F-score for each relation class as the evaluation metric. 
We tune the hyper-parameters such as the number and dimensions of layers and dropout rate on the development set using the hyper-parameter optimization framework Optuna~\cite{akiba-etal-2019-optuna} and the details are shown in \appendixref{sec:detail}. 
We employ the Adam~\cite{kingma-2015-adam} optimizer with the default parameters in PyTorch~\cite{NEURIPS2019_9015-paszke-pytorch} except for the learning rate. The training was performed without finetuning for the Longformer because the corpus is small to train a large transformer model.

We compare the following models on graphs initialized by the rule-based model (\textbf{with \rulebased{}}) and empty graphs (\textbf{without \rulebased{}}).\\
\textbf{\edit{}}: Proposed model\\
\textbf{\withoutitr{}}: \edit{} without iterative edge editing, i.e., $d_{max} = 1$.\\
\textbf{\withoutgcn{}}: \edit{} without GCN by replacing $\bar{\mathcal{N}}^{G}$ with $\bar{\mathcal{N}}$ in \eref{eq:edgeij}\\
\textbf{\randedit{}}: \edit{} with random-order editing\\
Additionally, we evaluate the following model with randomly initialized graphs.\\
\textbf{\randinit{}}: \edit{} with randomly connected edges, the number of which is same as that of the extraction results of \rulebased{}, with random classes 

Note that although we did not provide the direct comparison with the existing models, our \withoutgcn{} without \rulebased{} is similar to BRAN~\cite{verga-etal-2018-bran}; the only differences are that we use Longformer~\cite{Beltagy2020Longformer} instead of transformers, and NER training is not included. Moreover, most of the models for the document-level RE require dataset annotating both entities and their mentions, so the existing models like ATLOP~\cite{zhou-2021-atlop-relation-cdr-docred} cannot be directly applied to the current task.

\subsection{Results without \rulebased{}}
\label{ssec:res_edit_on_empty}

\begin{table}[t]
    \centering
    \begin{tabular}{lrr} \hline
        & \multicolumn{1}{c}{Dev} & \multicolumn{1}{c}{Test} \\ \hline
        \edit{} & \textbf{0.788}  & \textbf{0.729}\\
        \withoutitr{} & 0.732  & 0.685  \\
        \withoutgcn{} & 0.744  & 0.703  \\
        \randedit{} & 0.751 & 0.690 \\
        \randinit{} & 0.756 & 0.720  \\\hline
    \end{tabular}
    \caption{Evaluation results in micro F-score without \rulebased{}} 
    \label{tab:main_result_without_rule}
\end{table}

We show the results with empty initial graphs in \tref{tab:main_result_without_rule}. \edit{} shows the highest scores and this indicates the effectiveness of our approach when the initial graphs are empty. When we compare \edit{}, \withoutitr{}, and \randedit{}, we find that both iterative edge editing and close-first strategy are effective. Since \withoutgcn{} extracts from context without graph structure information, the better performance of \edit{} over \withoutgcn{} shows the effectiveness of the information in the graph structure. The low performance with \randinit{} shows that the edge information needs to be reliable.

\subsection{Results with \rulebased{}}
\label{ssec:res_edit}
\begin{table}[t]
    \centering
    \begin{tabular}{lrr} \hline
        & \multicolumn{1}{c}{Dev} & \multicolumn{1}{c}{Test} \\
        \hline
        \rulebased{} &0.797 & 0.807 \\\hline
        \edit{} & \textbf{0.878} & 0.851 \\
        \withoutitr{} & 0.863 & \textbf{0.863} \\
        \withoutgcn{} & 0.857  & 0.834  \\
        \randedit{} & 0.791 & 0.744  \\
        \hline
    \end{tabular}
    \caption{Evaluation results in micro F-score with \rulebased{}}
    \label{tab:main_result}
\end{table}

We summarize the results with \rulebased{} in \tref{tab:main_result}. We show the detailed results for \edit{} without \rulebased{}, \rulebased{}, and \withoutitr{} with \rulebased{} in \appendixref{sec:class-eval}.

When we compare the results with \tref{tab:main_result_without_rule}, the performance with \rulebased{} is better than the counterpart without \rulebased{} for all the settings. Furthermore, all the scores in \tref{tab:main_result} are better than those in \tref{tab:main_result_without_rule}, which shows the strength of \rulebased{}. 

Surprisingly, the results with our approach are better than that of \rulebased{} even though \rulebased{} is better than our approach without \rulebased{}. This indicates our \edit{} approach can make the prediction accurate. We can conclude that our \edit{} approach can utilize the information from the rule-based model and the initialization of the edges by \rulebased{} is useful.

As for the performance of the models, most results are consistent with \tref{tab:main_result_without_rule} except that \withoutitr{} shows the highest score on the test set. 
This may be partly because the initial graph by \rulebased{} is already reliable and editing does not help to improve the context. Results with \randedit{} support this since the performance degradation with \randedit{} is large compared to \tref{tab:main_result_without_rule} and \randedit{} is harmful in this case.
Moreover, the different behaviors on the development and test sets indicate an imbalance in the corpus split. 

\section{Case Study}

\begin{figure*}[t!]
    \centering
    \begin{tabular}{cccc}
    \multicolumn{4}{c}{
        \begin{minipage}[t]{\linewidth}
                \small
                \centering
        \fbox{\begin{minipage}{0.96\linewidth}
                    A series of \entity{polycrystalline} samples of \entity{SrMo1-xNixO4} (0.02$<$=x$<$=0.08) were \entity{prepared} through the conventional \entity{solid-state reaction method} in \entity{air}. \entity{Appropriate proportions} of \entity{high-purity} \entity{SrCO3}, \entity{MoO3}, and \entity{Ni} \entity{powders} were thoroughly \entity{mixed} according to the \entity{desired stoichiometry}, and then \entity{prefired} at \entity{900} [?]\entity{C} for \entity{24} \entity{h}. The \entity{obtained} \entity{powders} were \entity{ground}, \entity{pelletized}, and \entity{calcined} at \entity{1000}, \entity{1100} and \entity{1200} [?]\entity{C} for \entity{24} \entity{h} with intermediate grinding twice. White \entity{compounds}, \entity{SrMo1-xNixO4}, were \entity{obtained}. The \entity{compounds} were \entity{ground} and \entity{pressed} into small \entity{pellets} about \entity{10} \entity{mm} \entity{diameter} and \entity{2} \entity{mm} \entity{thickness}. These \entity{pellets} were \entity{reduced} in a \entity{H2/Ar} (\entity{5}\entity{\%}: \entity{95}\entity{\%}) flow at \entity{920} [?]\entity{C} for \entity{12} \entity{h}, and then the deep red colored products of \entity{SrMo1-xNixO3} were \entity{obtained}.
            \end{minipage}}
            \caption{Example document}
            \label{fig:text}
        \end{minipage}}  \\ \\
        \begin{minipage}[t]{0.21\linewidth}
            \centering
            \includegraphics[width=0.8\linewidth]{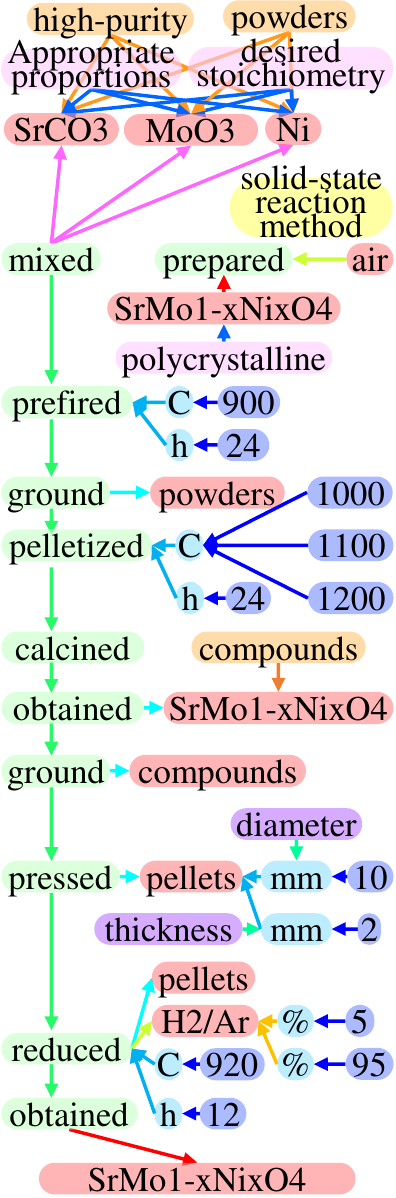}
            \caption{Gold graph for the document in \fref{fig:text}}
            \label{fig:gold}
        \end{minipage} &
        
        \begin{minipage}[t]{0.21\linewidth}
            \centering
            \includegraphics[width=0.8\linewidth]{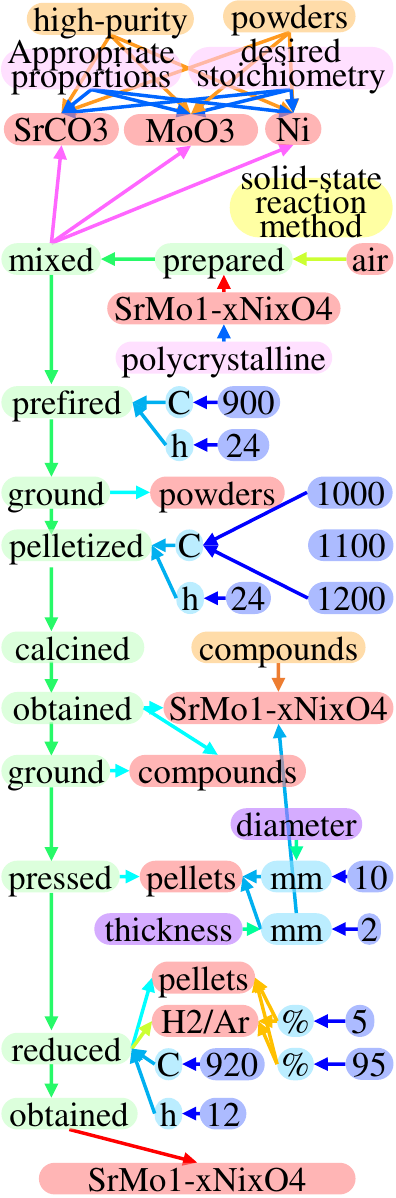}
            \caption{Example extraction results from the document in \fref{fig:text} by \edit{} without \rulebased{}}
            \label{fig:withoutrule}
        \end{minipage} &
        
        \begin{minipage}[t]{0.21\linewidth}
            \centering
            \includegraphics[width=0.8\linewidth]{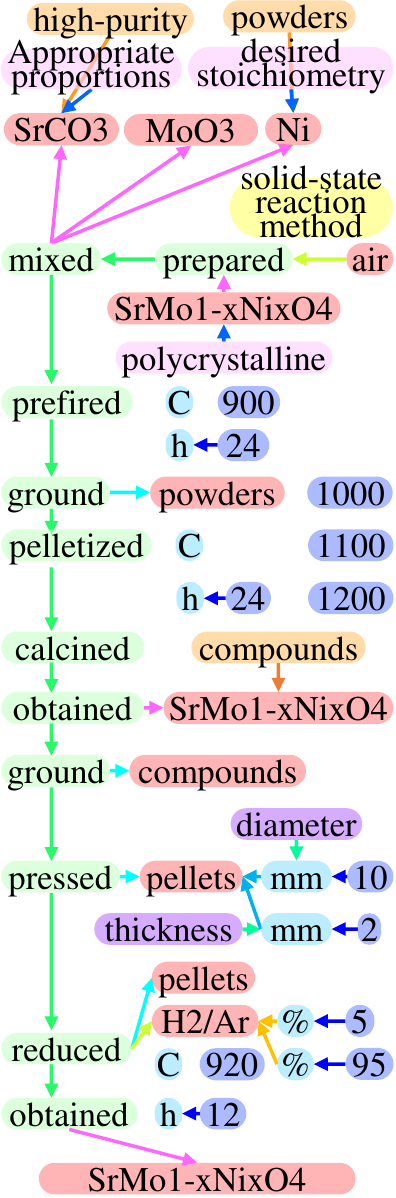}
            \caption{Example extraction results from the document in \fref{fig:text} by \rulebased{}}
            \label{fig:rule}
        \end{minipage} &
        \begin{minipage}[t]{0.21\linewidth}
            \centering
            \includegraphics[width=0.8\linewidth]{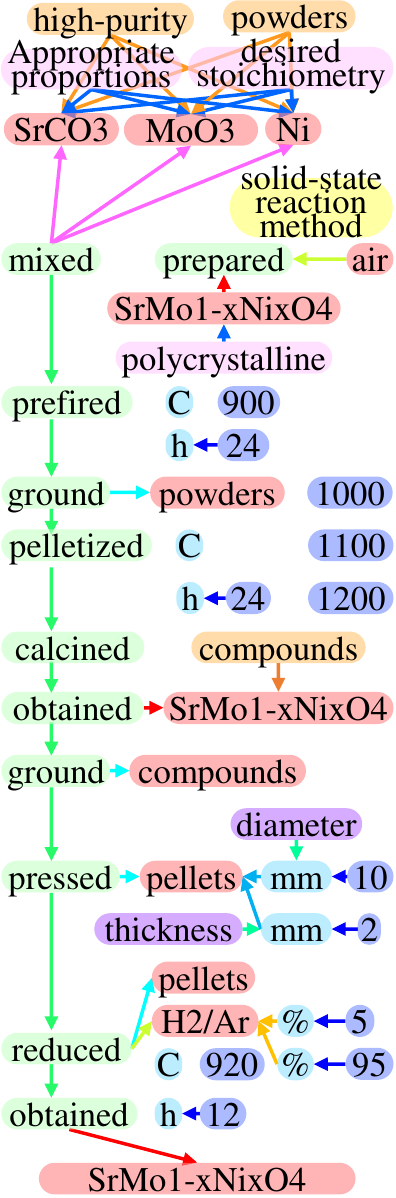}
            \caption{Example extraction results from the document in \fref{fig:text} by our \edit{} with \rulebased{}}
            \label{fig:withrule-edit}
        \end{minipage}
    \end{tabular}
\end{figure*}

We illustrated 6 graphs for an example document~\cite{zhang-et-al-2007-example-document} in the development data set shown in \fref{fig:text}: the result on the right side of \fref{fig:overview} shows our best extraction result using \withoutitr{} with \rulebased{}; \fref{fig:gold} shows the correct extraction; \fref{fig:withoutrule} shows the extraction result using \edit{} without \rulebased{}; \fref{fig:rule} shows the extraction result using \rulebased{}; and \fref{fig:withrule-edit} shows the extraction result using \edit{} with \rulebased{}. 
\fref{fig:gold} shows the material synthesis starts from \entity{mixed} with materials \entity{SrCO3}, \entity{MoO3} and \entity{Ni} to \entity{prefired} and so on, and the material \entity{SrMo1-xNixO4} is synthesized.
When we compare \fref{fig:withrule-edit} with \fref{fig:rule}, the extraction results are similar to \rulebased{}. Although the overall performance is low, \fref{fig:withoutrule}, which does not depend on the rule, extracts relations that are not extracted by the other systems and this shows the models with \rulebased{} and without \rulebased{} capture different relations.

\section{Related Work}
\label{sec:related-works}

RE has been widely studied to identify the relation between two entities in a sentence. In addition to traditional feature/kernel-based methods~\cite{zelenko2003kernel,miwa-sasaki-2014-modeling-joint-relation-close-first}, many neural RE methods have been proposed based on convolutional neural networks (CNNs)~\cite{zeng-etal-2014-relation}, recurrent neural networks (RNNs)~\cite{xu-etal-2015-classifying,miwa-bansal-2016-end}, graph convolutional networks (GCNs)~\cite{zhang-etal-2018-graph,schlichtkrull-2018-r-gcn}, and transformers~\cite{wang-etal-2019-extracting}. However, sentence-level RE is not enough to cover the relations in a document, and document-level RE has increasingly received research attention in recent years. 

Major approaches for document-level RE are graph-based methods and transformer-based methods. For graph-based methods, \citet{quirk-poon-2017-distant} first proposed a document graph for document-level RE. \citet{christopoulou-etal-2019-connecting} constructed a graph that included heterogeneous nodes such as entity mentions, entities, and sentences and represented edges between entities from the graph. \citet{nan-etal-2020-reasoning-lsr} proposed the automatic induction of a latent graph for relational reasoning across sentences.
The document graphs in these methods are defined on nodes of linguistic units such as words and sentences, which are different from our relation graphs. Unlike our method, these methods do not directly deal with relation graphs among entities.

For transformer-based methods, \citet{verga-etal-2018-bran} introduced a method to encode a document with transformers to obtain entity embedding and classify the relations between entities using the embedding. \citet{tang-etal-2020-relation-transformer} proposed a Hierarchical Inference Network (HIN) for document-level RE, which aggregates information from entity level to document level. \citet{zhou-2021-atlop-relation-cdr-docred} tackled document-level RE with an Adaptive Thresholding and Localized cOntext Pooling (ATLOP) model that introduces a learnable entity-dependent threshold for classification and aggregated local mention-level contexts that are relevant to both entities.

Several studies focus on procedural texts such as cooking recipes~\cite{bosselut2018simulating}, scientific processes~\cite{dalvi-etal-2018-tracking} and open domain procedures~\cite{tandon-etal-2020-dataset}. They, however, do not directly treat relation graphs.
Several efforts have been made to annotate procedural or action graphs in procedural text~\cite{mori2014flow,mysore-et-al-2019-olivetti-corpus,kuniyoshi-etal-2020-annotating}. \citet{kuniyoshi-etal-2020-annotating} and \citet{mehr2020universal} individually proposed rule-based systems to extract procedures from a document, but no neural methods have been proposed for the extraction.

\section{Conclusions}
We proposed a novel edge editing approach for document-level relation extraction. This approach treats the task as the edge editing of relation graphs, given nodes. It edits edges considering contexts in the document and the relation graph. 
We evaluated the approach on the material synthesis procedure corpus, and the results showed the usefulness of initializing edges by the rule-based model, utilizing prebuilt graph information for editing and editing in a close-first manner. As a result, our model performed an F-score of 86.3\% for edge prediction.

In future work, we plan to improve the approach to obtain more consistent and accurate relation graphs. We also would like to apply the approach to other data sets such as cooking recipes~\cite{mori2014flow} and temporal graphs~\cite{pustejovsky2003timebank,cassidy-etal-2014-annotation}. 

\bibliographystyle{acl_natbib}
\bibliography{ref}

\input{appendix}

\end{document}

%% file: appendix.tex
\def\nextope{\relationlabel{Next\_Operation}}
\def\propof{\relationlabel{Property\_Of}}
\def\numberof{\relationlabel{Number\_Of}}
\def\conditionof{\relationlabel{Condition\_Of}}
\def\amountof{\relationlabel{Amount\_Of}}
\def\descriptorof{\relationlabel{Descriptor\_Of}}
\def\apparatusof{\relationlabel{Apparatus\_Of}}
\def\typeof{\relationlabel{Type\_Of}}
\def\brandof{\relationlabel{Brand\_Of}}
\def\apparatusattrof{\relationlabel{Apparatus\_Attr\_Of}}
\def\recipeprecursor{\relationlabel{Recipe\_Precursor}}
\def\recipetarget{\relationlabel{Recipe\_Target}}
\def\participantmat{\relationlabel{Participant\_Material}}
\def\solventmat{\relationlabel{Solvent\_Material}}
\def\atmosphericmat{\relationlabel{Atmospheric\_Material}}
\def\corefof{\relationlabel{Coref\_Of}}

\def\ope{\entitylabel{Operation}}
\def\mat{\entitylabel{Material}}
\def\propmisc{\entitylabel{Property-Misc}}
\def\nonmat{\entitylabel{Nonrecipe-Material}}
\def\propunit{\entitylabel{Property-Unit}}
\def\num{\entitylabel{Number}}
\def\amountunit{\entitylabel{Amount-Unit}}
\def\conditionunit{\entitylabel{Condition-Unit}}
\def\conditionmisc{\entitylabel{Condition-Misc}}
\def\amountmisc{\entitylabel{Amount-Misc}}
\def\matdescriptor{\entitylabel{Material-Descriptor}}
\def\apparatusdescriptor{\entitylabel{Apparatus-Descriptor}}
\def\apparatus{\entitylabel{Apparatus}}
\def\synthapparatus{\entitylabel{Synthesis-Apparatus}}
\def\charaapparatus{\entitylabel{Characterization-Apparatus}}
\def\proptype{\entitylabel{Property-Type}}
\def\conditiontype{\entitylabel{Condition-Type}}
\def\apparatusproptype{\entitylabel{Apparatus-Property-Type}}
\def\apparatusunit{\entitylabel{Apparatus-Unit}}
\def\brand{\entitylabel{Brand}}
\def\meta{\entitylabel{Meta}}
\def\r{\entitylabel{Reference}}
\setlength\titlebox{2cm}

\appendix

\section{Statistics of the Materials Science Procedural Text Corpus}
\label{sec:stat-corpus}

We present the statistics of the materials science procedural text corpus\footnote{\url{https://github.com/olivettigroup/annotated-materials-syntheses}} proposed by \citet{mysore-et-al-2019-olivetti-corpus}. \tref{tab:num_entity} and \tref{tab:num_relation} summarize the numbers of entities and relations, respectively.

\section{Rule-based Relation Extraction Model}
\label{sec:rule}

We built a rule-based model by defining the rules to extract relations between entity pairs for the materials science procedural text corpus~\cite{mysore-et-al-2019-olivetti-corpus}. The rules were adapted from the rule-based model in \citep{kuniyoshi-etal-2020-annotating} for the target corpus. The rules depend on labels of the entities of an entity pair, distance, and the order of occurrence of the entities. According to the combination of labels of the entities, our rules are divided into three types: \ope{}--\ope{}, \ope{}--\mat{} and other relations. In the following, the starting point of a relation is called head and the ending point is called tail, and an edge is denoted as \entitylabel{Head}--\entitylabel{Tail}.

\subsection{\ope{}--\ope{}}
The relation \ope{}--\ope{} takes only a \nextope{} label, which means the progress of operation.

\nextope{}: 
Close \ope{} entities are linked with the relation from the beginning to the end in the document order, in which the entities of \ope{} appear.

\subsection{\ope{}--\mat{}}
For the edges of \ope{}--\mat{}, there are five relation labels: \recipeprecursor{} indicates the input of a material; \recipetarget{} indicates the generation of a product; \participantmat{} indicates the generation of an intermediate product; \solventmat{} indicates the solvent material of an operation; and \atmosphericmat{}  indicates the atmosphere of an operation.

\begin{table}[t!]
    \centering
    \small
    \begin{tabular}{lrrr} \hline
        Entity class & Train & Dev & Test \\ \hline
        \mat{} & 4,271 & 277 & 316 \\ 
        \ope & 3,249 & 212 & 242 \\ 
        \num{} & 2,872 & 224 & 219 \\ 
        \conditionunit{} & 1,363 & 101 & 87 \\ 
        \matdescriptor{} & 1,214 & 67 & 89 \\ 
        \amountunit{} & 1,193 & 96 & 98 \\ 
        \propmisc{} & 481 & 25 & 16 \\ 
        \conditionmisc{} & 468 & 32 & 20 \\ 
        \synthapparatus{} & 433 & 20 & 34 \\ 
        \nonmat{} & 329 & 33 & 25 \\ 
        \brand{} & 291 & 30 & 27 \\ 
        \apparatusdescriptor{} & 165 & 10 & 9 \\ 
        \amountmisc{} & 149 & 14 & 7 \\ 
        \meta{} & 128 & 12 & 13 \\ 
        \proptype{} & 124 & 10 & 4 \\ 
        \conditiontype{} & 119 & 2 & 1 \\ 
        \r{} & 106 & 10 & 11 \\ 
        \propunit{} & 92 & 7 & 8 \\ 
        \apparatusunit{} & 89 & 6 & 16 \\ 
        \entitylabel{Character.-Apparatus} & 54 & 2 & 11 \\ 
        \apparatusproptype{} & 26 & 0 & 6 \\ 
        \hline
    \end{tabular}
    \caption{Entities in the materials science procedural text corpus}
    \label{tab:num_entity}
\end{table}

\begin{table}[t!]
    \centering
    \small
    \begin{tabular}{lrrr} \hline
        Relation class & train & dev & test \\ \hline
        \nextope{} & 2,898 & 184 & 202\\ 
        \recipeprecursor{} & 876 & 67 & 89\\ 
        \recipetarget{} & 363 & 31 & 22\\ 
        \participantmat{} & 1,723 & 113 & 124\\ 
        \solventmat{} & 463 & 28 & 33\\ 
        \atmosphericmat{} & 183 & 11 & 14\\ 
        \propof{} & 586 & 35 & 21\\ 
        \conditionof{} & 1,810 & 132 & 107\\ 
        \numberof{} & 2,805 & 219 & 209\\ 
        \amountof{} & 1,512 & 130 & 121\\ 
        \descriptorof{} & 1,495 & 91 & 102\\ 
        \brandof{} & 423 & 42 & 41\\ 
        \typeof{} & 164 & 7 & 13\\ 
        \apparatusof{} & 455 & 20 & 36\\ 
        \apparatusattrof{} & 90 & 6 & 11 \\
        \corefof{} & 267 & 12 & 14\\ 
        \hline
    \end{tabular}
    \caption{Relations in the materials science procedural text corpus}
    \label{tab:num_relation}
\end{table}

For \solventmat{}, \atmosphericmat{} and \participantmat{} labels, a dictionary is prepared manually for each label. The relations are linked from the nearest \ope{} to a \mat{} in the sentence if the \mat{} match in the dictionary since these relations take specific \mat{} entities. The dictionary is included in the source code.

\recipeprecursor{} is linked from all \mat{} that do not match the dictionary of \solventmat{}, \atmosphericmat{}, and \participantmat{} to the nearest \ope{}. This rule-based model does not produce the relation \recipetarget{}. The  reason for these decisions is that it is difficult to classify these relations with simple rules.

\subsection{Remaining Relations}

The remaining 9 relation labels are defined between the other pairs of entity labels: \propof{}, which indicates a condition of a material; \conditionof{}, which indicates a condition of an operation; \numberof{}, which indicates the relationship between a number and a unit; \amountof{}, which indicates a condition of a quantity; \typeof{}, which indicates a condition of a numerical condition; \brandof{}, which indicates the brand of a material or equipment; \apparatusof{}, which indicates equipment used in an operation; \apparatusattrof{}, which indicates a numerical condition of on equipment; and \descriptorof{}, which indicates other conditions. For these labels, the rules are defined based only on the labels of head and tail entities and the distance between them. We explain the detailed rules in the remainder of this section.

\propof{}: The relation can take \propunit{} or \propmisc{} as the head and \mat{} or \nonmat{} as the tail. When \propunit{} is a head, it is linked with the nearest \mat{} in the sentence. When \propmisc{} is a head, it is linked to the nearest \mat{} or \nonmat{} in the sentence. 

\conditionof{}: \conditionunit{} and \conditionmisc{} are linked to the nearest \ope{} with the relation in the sentence.

\numberof{}: \num{} is linked to the nearest \propunit{}, \conditionunit{}, or \apparatusunit{} that appear after the \num{} in the sentence.

\amountof{}: The relation is linked from \amountunit{} and \amountunit{} to the nearest \mat{} or \nonmat{} in the sentence.

\descriptorof{}: When \matdescriptor{} is a head, it is linked to the nearest \mat{} or \nonmat{} in the sentence.
When \apparatusdescriptor{} is a head, it is linked to the nearest \synthapparatus{} in the sentence.

\apparatusof{}: The relation is linked from \synthapparatus{} and \charaapparatus{} to the nearest \ope{} with the priority given to the \ope{} that appear before the \apparatus{} in the sentence.  

\typeof{}: \proptype{} and \apparatusproptype{} are linked to the nearest \propunit{} and \apparatusunit{} in the sentence with the relation, respectively. When \conditiontype{} is a head, it is linked to the nearest \conditionunit{} that appears before the \conditiontype{} in the sentence.

\brandof{}: The relation is linked from \brand{} to the nearest entities that may have brands (i.e., \mat{}, \nonmat{}, \synthapparatus{}, and \charaapparatus{}) in the sentence. 

\apparatusattrof{}: \apparatusunit{} is linked to the nearest \synthapparatus{} or \charaapparatus{}. 

\corefof{}: The relation is not detected by the rules because it is difficult to describe rules. 

\begin{table}[t!]
    \centering
    \small
    \begin{tabular}{lcccr} \hline
        Parameter & Range & Value \\ \hline
        Learning rate & [1e-5, 1e-2) & 0.001\\
        No. of GCN layers &[0, 4] & 3 \\
        $d_{max}$ & [1, 10] & 4 \\
        Dimension of hidden layers & [32, 128] & 85 \\
        No. of $\mathrm{FC}^{out}$ layers & [1, 5] & 4\\
        No. of $\mathrm{FC}^{h}$ and $\mathrm{FC}^{t}$ layers &  [1, 5] & 1\\
        Dropout rate & [0.0, 1.0) & 0.46\\
        Dimension of $\bm e_{ij}^{old}$ & [1, 32] & 3\\
        Maximum distance for $\bm b_{ij}$ & [1, 32] & 3\\
        Dimension of $\bm b_{ij}$ & [1, 100] & 1\\
        Use bidirectional GCN & True or False & True \\ \hline
    \end{tabular}
    \caption{Search space for optimization of hyper-parameters and the selected values after optimization}
    \label{tab:search-space}
\end{table}

\section{Tuning Details}
\label{sec:detail}
\begin{table}[t!]
    \small
    \centering
    \begin{tabular}{lrrr} \hline
         Relation & Prec. & Recall & F-score \\ \hline
        \nextope{} & 0.622 & 0.693 & 0.656 \\
        \recipeprecursor{} & 0.632 & 0.539 & 0.582 \\
        \recipetarget{} & 0.640 & 0.727 & 0.681 \\
        \participantmat{} & 0.641 & 0.476 & 0.546 \\
        \solventmat{} & 0.491 & 0.818 & 0.614 \\
        \atmosphericmat{} & 0.733 & 0.786 & 0.759 \\
        \propof{} & 0.773 & 0.810 & 0.791 \\
        \conditionof{} & 0.798 & 0.850 & 0.824 \\
        \numberof{} & 0.874 & 0.962 & 0.916 \\
        \amountof{} & 0.722 & 0.645 & 0.681 \\
        \descriptorof{} & 0.761 & 0.814 & 0.787 \\
        \brandof{} & 0.567 & 0.415 & 0.479 \\
        \typeof{} & 0.900 & 0.692 & 0.783 \\
        \apparatusof{} & 0.657 & 0.639 & 0.648 \\
        \apparatusattrof{} & 0.769 & 0.909 & 0.833 \\
        \corefof{} & 0.875 & 0.500 & 0.636 \\\hline
        Overall & 0.717 & 0.722 & 0.720 \\\hline
    \end{tabular}
    \caption{Detailed results using \edit{} without \rulebased{} on the test set }
    \label{tab:edit-without-rule-class-eval}
\end{table}

\begin{table}[t!]
    \small
    \centering
    \begin{tabular}{lrrr} \hline
         Relation & Prec. & Recall & F-score \\ \hline
        \nextope{} & 0.990 & 0.881 & 0.932 \\
        \recipeprecursor{} & 0.730 & 0.414 & 0.528\\
        \recipetarget{} & 0.000 & 0.000 & 0.000\\
        \participantmat{} & 0.419 & 0.800 & 0.550\\
        \solventmat{}  & 0.697 & 0.418 & 0.522\\
        \atmosphericmat{} & 1.000 & 0.378 & 0.549\\
        \propof{} & 0.905 & 1.000 & 0.950\\
        \conditionof{} & 0.963 & 0.981 & 0.972 \\
        \numberof{} & 0.943 & 0.961 & 0.952\\
        \amountof{} & 0.744 & 0.865 & 0.800 \\
        \descriptorof{} & 0.931 & 0.979 & 0.955 \\
        \brandof{} & 0.561 & 0.920 & 0.697 \\
        \typeof{} & 0.769 & 1.000 & 0.870\\
        \apparatusof{} & 0.972 & 0.854 & 0.909\\
        \apparatusattrof{} & 0.909 & 0.769 & 0.833\\
        \corefof{} &  0.000 & 0.000 & 0.000 \\ \hline
        Overall & 0.807 & 0.808 & 0.807 \\ \hline
    \end{tabular}
    \caption{Detailed results with \rulebased{} on the test set}
    \label{tab:rule-class-eval}
\end{table}

\begin{table}[t!]
    \small
    \centering
    \begin{tabular}{lrrr} \hline
         Relation & Prec. & Recall & F-score \\ \hline
        \nextope{} & 0.905 & 0.990 & 0.946 \\
        \recipeprecursor{} & 0.810 & 0.573 & 0.671 \\
        \recipetarget{} & 0.560 & 0.636 & 0.596 \\
        \solventmat{} & 0.733 & 0.667 & 0.698 \\
        \participantmat{} & 0.624 & 0.790 & 0.698 \\
        \atmosphericmat{} & 0.778 & 1.000 & 0.875 \\
        \propof{} & 0.905 & 0.905 & 0.905 \\
        \conditionof{} & 0.953 & 0.944 & 0.948 \\
        \numberof{} & 0.958 & 0.990 & 0.974 \\
        \amountof{} & 0.854 & 0.868 & 0.861 \\
        \descriptorof{} & 0.941 & 0.931 & 0.936 \\
        \brandof{} & 0.880 & 0.537 & 0.667 \\
        \typeof{} & 1.000 & 0.692 & 0.818 \\
        \apparatusof{} & 0.833 & 0.972 & 0.897 \\
        \apparatusattrof{} & 0.769 & 0.909 & 0.833 \\
        \corefof{} & 0.750 & 0.429 & 0.545 \\\hline
        Overall & 0.856 & 0.870 & 0.863 \\\hline
    \end{tabular}
    \caption{Detailed results using \withoutitr{} with \rulebased{} on the test set}
    \label{tab:ie-with-rule-class-eval}
\end{table}

We tuned our model using a hyper-parameter optimization framework Optuna~\cite{akiba-etal-2019-optuna}. We searched for the hyper-parameters that maximize micro-F scores within 600 trials on the development set. We employed the tree-structured Parzen estimator algorithm~\cite{bergstra-2011-tpe-sampler} for the sampler and the successive halving algorithm~\cite{li-etal-2020-successive-halving} for the pruner with default options in Optuna. In each trial of the search, we trained our model for 100 epochs, which was confirmed by preliminary experiments to be sufficient for convergence. We searched hyper-parameters on 20 NVIDIA GPUs, which include Tesla V100, TITAN V, RTX 3090, and GTX TITAN Xp GPUs.

We defined the search space as shown in \tref{tab:search-space}; the hyper-parameters for the search are composed of the learning rate for Adam, the number of GCN layers, the maximum edit distance $d_{max}$, the dimensions of all hidden layers, the number of $\mathrm{FC}^{out}$ layers, the number of $\mathrm{FC}^{h}$ and $\mathrm{FC}^{t}$ layers, the dropout rate, the dimension of $\bm e_{ij}^{old}$, the maximum distance and the dimension for $\bm b_{ij}$ and whether to use bidirectional GCNs or uni-directional GCNs. 
In the table, the range column shows the range of values to search and the final value column shows the rounded selected values after the optimization.

\section{Detailed Evaluation Results}
\label{sec:class-eval}
Our editing models for evaluation are trained with a TITAN V GPU for \edit{} with \rulebased{} and a Tesla V100 GPU for the others. The training takes about 6 hours 30 minutes with \withoutitr{} using \rulebased{} and 21 hours with \edit{} not using \rulebased{}. 

We show the detailed evaluation results with precision (Prec.), recall, and F-score on the test set in \tref{tab:edit-without-rule-class-eval} for \edit{} without \rulebased{}, \tref{tab:rule-class-eval} for \rulebased{}, and \tref{tab:ie-with-rule-class-eval} for \withoutitr{} without \rulebased{}. 
The results show the relations that are not covered by \rulebased{}, i.e., \recipetarget{} and \corefof{}, are extracted by our approach, and for these classes, \edit{} without \rulebased{} show the better performance than the models with \rulebased{}. 
Some relations with high performance by \rulebased{}, including \nextope{}, \conditionof{}, and \descriptorof{}, are extracted by \withoutitr{} with \rulebased{} in high performance. This shows our approach can effectively utilize the outputs of \rulebased{}.